\documentclass{article}

\usepackage{microtype}
\usepackage{graphicx}
\usepackage{subfigure}
\usepackage{booktabs} 
\usepackage{tcolorbox}
\usepackage[utf8]{inputenc} 
\usepackage[T1]{fontenc}    
\usepackage{hyperref}       
\usepackage{url}            

\usepackage{amsfonts}       
\usepackage{nicefrac}       
\usepackage{microtype}      

\usepackage{amsmath}
\usepackage{amssymb}
\usepackage{placeins}

\DeclareMathOperator*{\argmax}{arg\,max}

\usepackage{hyperref}

\usepackage[accepted]{icml2020}

\icmltitlerunning{GLAD: {\bf GL}ocalized {\bf A}nomaly {\bf D}etection via Human-in-the-Loop Learning}

\begin{document}

\twocolumn[
\icmltitle{GLAD: {\bf GL}ocalized {\bf A}nomaly {\bf D}etection via Human-in-the-Loop Learning}



\icmlsetsymbol{equal}{*}

\begin{icmlauthorlist}
\icmlauthor{Md Rakibul Islam}{goo}
\icmlauthor{Shubhomoy Das}{goo}
\icmlauthor{Janardhan Rao Doppa}{goo}
\icmlauthor{Sriraam Natarajan}{good}
\end{icmlauthorlist}

\icmlaffiliation{goo}{School of EECS, Washington State University, Pullman, WA 99163}
\icmlaffiliation{good}{School of Engineering \& Computer Science,
The University of Texas at Dallas, TX 75080}

\icmlcorrespondingauthor{Md Rakibul Islam}{mdrakibul.islam@wsu.edu}
\icmlcorrespondingauthor{Shubhomoy Das}{shubhomoy.das@wsu.edu}
\icmlcorrespondingauthor{Janardhan Rao Doppa}{jana.doppa@wsu.edu}
\icmlcorrespondingauthor{Sriraam Natarajan}{Sriraam.Natarajan@utdallas.edu}

\icmlkeywords{Machine Learning, ICML}

\vskip 0.3in
]



\printAffiliationsAndNotice{\icmlEqualContribution from first two authors.} 

\begin{abstract}

{
	Human analysts that use anomaly detection systems in practice want to retain the use of simple and explainable {\em global} anomaly detectors. In this paper, we propose a novel human-in-the-loop learning algorithm called GLAD (GLocalized Anomaly Detection) that supports {\em global} anomaly detectors. GLAD automatically learns their {\em local} relevance to specific data instances using label feedback from human analysts. The key idea is to place a uniform prior on the relevance of each member of the anomaly detection ensemble over the input feature space via a neural network trained on unlabeled instances. Subsequently, weights of the neural network are tuned to adjust the local relevance of each ensemble member using all labeled instances. GLAD also provides explanations which can improve the understanding of end-users about anomalies. Our experiments on synthetic and real-world data show the effectiveness of GLAD in learning the local relevance of ensemble members and discovering anomalies via label feedback.}
\end{abstract}

\section{Introduction}

\paragraph{Definition 1 (\textit{Glocal})}  \textit{Reflecting or characterized by both local and global considerations\footnote{\scriptsize{\url{https://en.wikipedia.org/wiki/Glocal} (retrieved on May-21-2020)}}.}

End-users find it easier to trust algorithms they understand and are familiar with. Such algorithms are typically built on broadly general and simplifying assumptions over the entire feature space (i.e., {\em global} behavior), which may not be applicable universally (i.e., not relevant {\em locally} in some parts of the feature space) in an application domain. This observation is true of most machine learning algorithms including those for anomaly detection. We propose a principled technique referred as \textit{GLocalized Anomaly Detection (GLAD)} which allows a human analyst to continue using anomaly detection ensembles with global behavior by learning their local relevance in different parts of the feature space via label feedback. 

Ensembles of anomaly detectors often outperform single detectors \cite{aggarwal:2017}. Additionally, anomalous instances can be discovered faster when the ensembles are used in conjunction with active learning, where a human analyst labels the queried instance(s) as {\em nominal} or {\em anomaly} \cite{veeramachaneni:2016,das:2016,das:2018,siddiqui:2018}. A majority of the active learning techniques for discovering anomalies employ a weighted linear combination of the anomaly scores from the ensemble members. This approach works well when the members are themselves highly localized, such as the leaf nodes of tree-based detectors \cite{das:2018}. However, when the members of the ensemble are global (such as LODA projections \cite{pevny:2015}), it is highly likely that individual detectors are incorrect in at least some local parts of the input feature space. 

To overcome this drawback, our GLAD algorithm automatically learns the {\em local relevance} of each ensemble member in the feature space via a neural network using the label feedback from a human analyst. One interesting observation related to the key insight behind active learning with tree-based models (Tree-AAD) \cite{das:2018} and GLAD is as follows: uniform prior over weights of each subspace (leaf node) in Tree-AAD and uniform prior over input feature space for the relevance of each ensemble member in GLAD are highly beneficial for label-efficient active learning. We can consider GLAD as very similar to the Tree-AAD approach. Tree-AAD partitions the input feature space into discrete subspaces and then places a uniform prior over those subspaces (i.e., the uniform weight vector to combine ensemble scores). If we take this view to an extreme by imagining that each instance in feature space represents a subspace, we can see the connection to GLAD. While Tree-AAD assigns the scores of discrete subspaces to instances (e.g., node depths for Isolation Forest), the scores assigned by GLAD are continuous, defined by the global ensemble members. The {\em relevance} in GLAD is analogous to the {\em learned weights} in Tree-AAD.

Our GLAD technique is similar in spirit to dynamic ensemble weighting \cite{jimenez:1998}. However, since we are in an active learning setting for anomaly detection, we need to consider two important aspects: {\bf (a)} Number of labeled examples is very small (possibly none), and {\bf (b)} To reduce the effort of the human analyst, the algorithm needs to be \textit{primed} so that the likelihood of discovering anomalies is very high from the first feedback iteration itself. Specifically, we employ a neural network to predict the local relevance of each ensemble member. This network is primed with unlabeled data such that it places a uniform prior for the relevance of each ensemble member over the input feature space. In each iteration of the active learning loop, we select one unlabeled instance for querying, and update the weights of the neural network to adjust the local relevance of each ensemble member based on all the labeled instances. Our code and datasets are publicly available at  \url{https://github.com/shubhomoydas/ad_examples}.

\section{Related Work}
\label{sec:related}

\textbf{Anomaly detection} approaches are mostly unsupervised. It assumes that the concept of nominal and anomaly can be derived from the dataset. \cite{scholkopf2001estimating, breunig:00,liu:08,pevny:2015,emmott:2015} are some of the classical algorithms for anomaly detection. One major problem of such approaches is the were not designed to incorporate feedbacks. And that introduced a lot of false alarms from the model. Inherent bias was the problem for such high false alarms. Ensemble based approaches were proposed to improve the performance \cite{aggarwal:2017}. Some other variants are heterogeneous detectors \cite{senator:2013}, GMM \cite{emmott:2015}. The state of the art anomaly detection approach \cite{liu:08} is also an ensemble based approach.

For supervised and semi-supervised anomaly detection main assumptions are the presence of labels. And semi-supervised approaches like \cite{munoz2010semisupervised, blanchard2010semi} was developed on an assumption that labels for nominals are only present. Later, \cite{Grnitz2013TowardSA} considers a semi-supervised algorithm for anomaly detection and employs active learning. Besides, there were some clustering assumptions made at \cite{zhu2005semi, chapelle2009semi} for semi supervised settings. This assumption breaks when the problems is being applied for anomaly detection as the anomalies usually do not produce clusters in the data space. \\

\textbf{Active learning} based approaches for anomaly detection is becoming an important research area \cite{das:2017,siddiqui:2018,TKDD-2020,veeramachaneni:2016,das:2016,guha:2016,nissim:14,stokes:2008,he:2008,almgren:2004,abe:2006}. To deploy anomaly detection systems in real-world this is a necessity for end user. It enables domain experts to interact with the system and update the model.

\textbf{Explainability} is an essential component for any learning based model \cite{doshi-velez:2017}. The main objective of explainability is to help humans(end-users) understanding about the model and tools. Previous studies focused on \textit{ruleset} based also known as disjunctive normal form (DNF) based explanatations. Simplicity of such models made them easily accessible for humans \cite{letham:2015,goh:2014, frnkranz:2012}. Another direction for explainabiltiy is to develop a model-agnostic mechanism. Some notable works are {\em LIME} \cite{ribeiro:2016}, {\em Anchors} \cite{ribeiro:2018}, and {\em x-PACS} \cite{macha:2018} where they provide explanations for any pretrained model. For GLAD, model-agnostic techniques can be applied for generic ensembles. GLAD first identifies the most relevant ensemble member for an anomaly instance. Subsequently, the model-agnostic techniques can be employed to explain or describe the predictions of that detector.

\section{Problem Setup}
\label{sec:problem}

We will denote the input feature space by $\mathcal{X} \subseteq \mathbb{R}^d$. We are given a dataset ${\bf D} = \{{\bf x}_1, ..., {\bf x}_n\}$, where ${\bf x}_i \in \mathcal{X}$ is a data instance that is associated with a hidden label $y_i \in \{-1, +1\}$. Instances labeled $+1$ represent the \textit{anomaly} class and are at most a small fraction $\tau$ of all instances. The label $-1$ represents the \textit{nominal} class. We also assume the availability of an ensemble $\mathcal{E}$ of $M$ {\em global} anomaly detectors (e.g., LODA projections) which assign scores $s_1({\bf x}), s_2({\bf x}),\cdots, s_M({\bf x})$ to each instance ${\bf x} \in \mathcal{X}$ such that instances labeled $+1$ tend to have scores higher than instances labeled $-1$. Suppose that $p_m({\bf x}) \in [0, 1]$ denotes the relevance of the $m^{th}$ ensemble member (via a neural network) for a data instance ${\bf x}$. We combine the scores of $M$ anomaly detectors as follows: \texttt{Score}(${\bf x}$) = $\sum_{m=1}^M{s_m({\bf x}) \cdot p_m({\bf x})}$. Our human-in-the-loop learning algorithm assumes the availability of a human analyst who can provide the true label for any instance. The overall goal is to learn the local relevance of ensemble members (i.e., appropriate weights of the neural network) for maximizing the number of true anomalies shown to the human analyst.
\begin{figure}[h]
	\centering
	\includegraphics[width=0.5\textwidth]{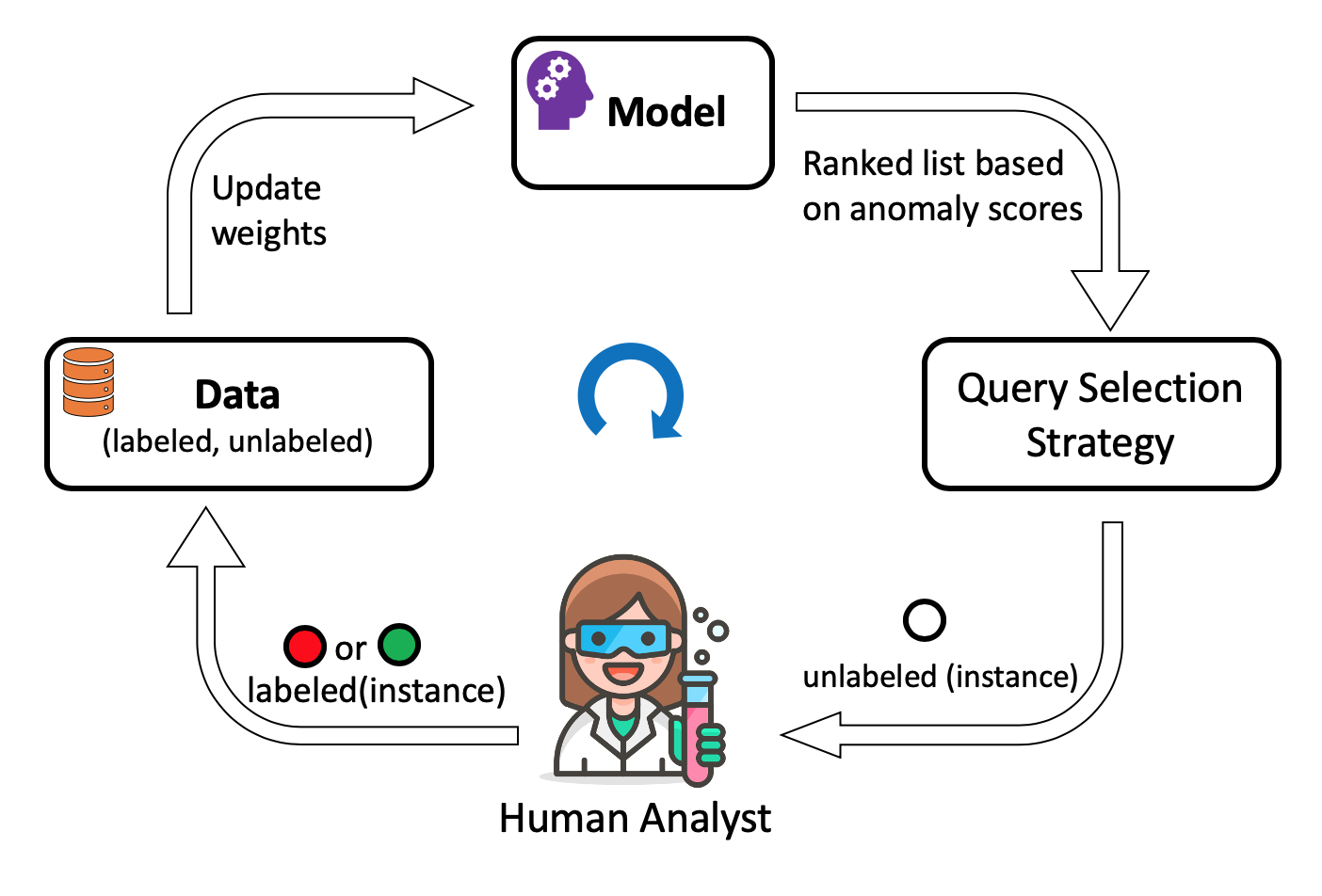} \\
	\caption{High-level overview of the human-in-the-loop anomaly detection framework. Our \textit{GLAD} algorithm instantiates the ``Model'' component by placing a uniform prior over the input feature space using a neural network.}
	\label{fig:framework}
\end{figure}




\section{GLAD Algorithm}
\label{sec:algorithm}

\begin{figure*}[htb]
	\centering
	\includegraphics[width=.95\textwidth]{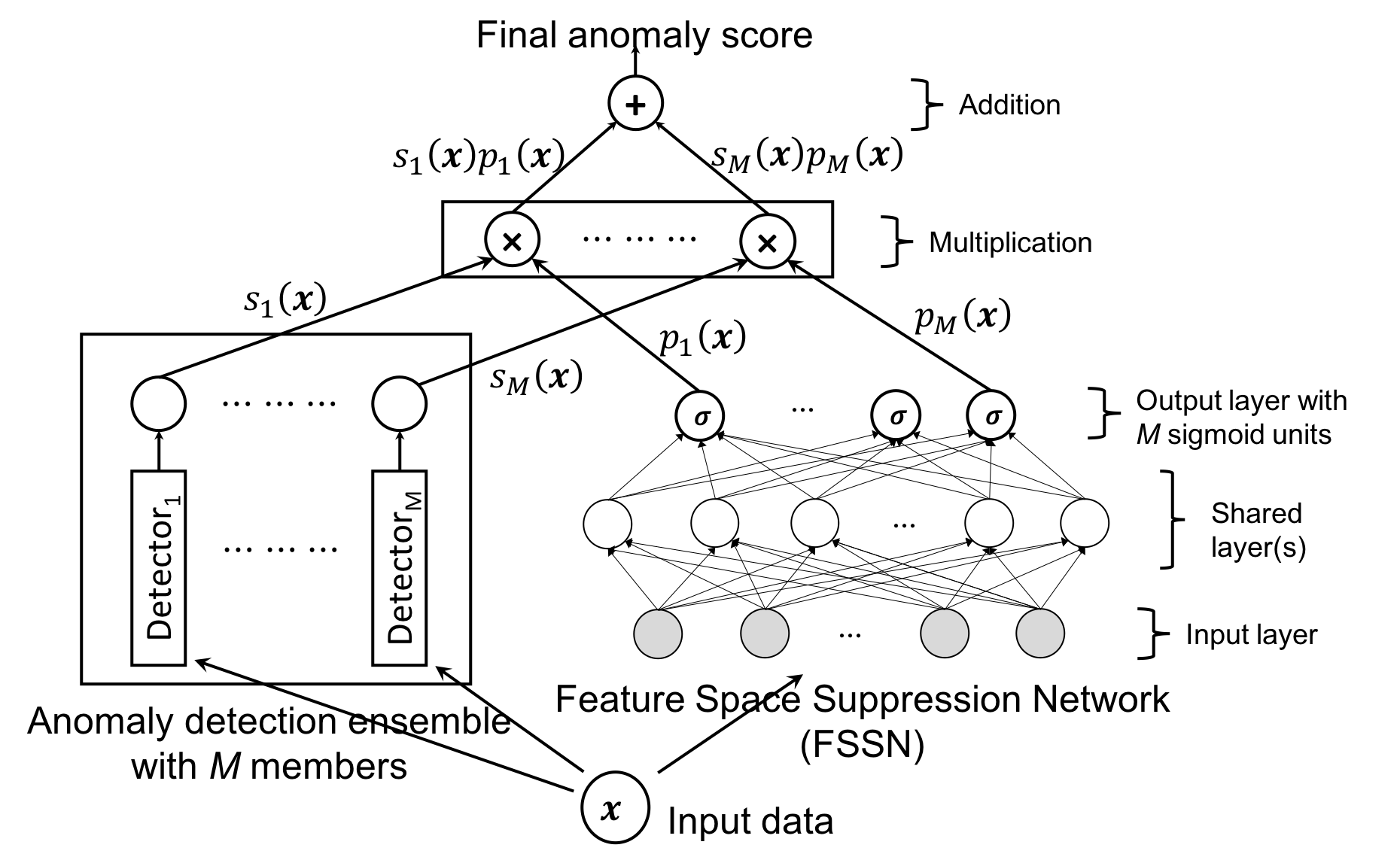} \\
	\caption{Overview of the model component of GLAD algorithm. The anomaly detection ensemble contains $M$ global detectors. We assume that all ensemble members are pre-trained and cannot be modified. The final layer of the Feature Space Suppression Network (FSSN) contains $M$ sigmoid outputs, each one paired with a corresponding ensemble member. Each output node in the FSSN is initially primed to predict the same probability ($0.5$ in our experiments) across the entire input feature space. FSSN learns which parts of the feature space are \textit{\textbf{relevant}} for each detector based on the label feedback received from human analyst. For a given data instance ${\bf x}$, $s_m({\bf x})$ denotes the score assigned to it by the $m^{th}$ detector and $p_m({\bf x})$ denotes the probability computed by the FSSN that the $m^{th}$ detector is relevant. The final anomaly score for data instance ${\bf x}$ is the sum of all scores from each detector weighted by their corresponding relevance. In each iteration of the active learning loop, we select one unlabeled instance for querying, and update the weights of FSSN to adjust the local relevance of each ensemble member over all labeled instances.}
	\label{fig:architecture}
\end{figure*}
\paragraph{Overview.} We start with the assumption that each ensemble member is uniformly relevant in every part of the input feature space. This assumption is implemented by priming a neural network referred to as \textit{FSSN} (feature space suppression network) to predict the same probability value $b \in (0, 1)$ for every instance in ${\bf D}$. In effect, this mechanism places \textit{\textbf{a uniform prior over the input feature space $\mathcal{X}$} for the relevance of each detector}. Subsequently, the algorithm receives label feedback from a human analyst and determines whether the ensemble made an error (i.e., anomalous instances are ranked at the top and scores of anomalies are higher than scores of nominals). If there is an error, the weights of FSSN are updated to suppress all erroneous detectors for similar inputs in the future. Figure~\ref{fig:architecture} illustrates different components of the GLAD model including the ensemble of anomaly detectors and the Feature Space Suppression Network (FSSN). And algorithm ~\ref{alg:hill} illustrates how the GLAD model fits inside the overall human-in-the loop framework. The GLAD components are highlighted inside the framework.


\paragraph{AAD Loss.} We employ the AAD hinge loss from \cite{das:2018} to measure the degree of error in anomaly detection based on all labeled instances. This loss is a simplified version of the constraint-based loss proposed in \cite{das:2016}, and is more suitable for gradient-based learning. AAD makes two assumptions: {\bf (a)} $\tau$ fraction of instances (a very small number of instances from ${\bf D}$) are anomalous, and {\bf (b)} labeled anomalies should have scores higher than the instance currently ranked at the $\tau$-th quantile,  whereas nominals should have scores lower than that instance. We will denote this loss by $\ell_{AAD}({\bf x})$.

\begin{algorithm}[t]
	\caption{\texttt{GLAD} ($B$, $\mathcal{E}$, {\em FSSN}, ${\mathbf D}$, ${\mathbf H}_f$, $b$)}
	\label{alg:hill}
	\begin{algorithmic}
		\STATE \textbf{Input:} Query budget $B$, Ensemble of global anomaly detectors $\mathcal{E}$, FSSN neural network with parameters ${\bf \Theta}$, complete dataset ${\mathbf D}$, labeled instances ${\mathbf H}_f \subseteq {\bf D}$, bias probability $b$ 
		\STATE {\bf Priming Step:} Initialize {\em FSSN} to predict $b$ for all ${\bf x} \in {\bf D}$
		\FOR{$t \in \{1 \cdots B\}$}
    		\STATE
    		\begin{tcolorbox}[colback=green!5!white,colframe=green!75!black]
    		\STATE // Score unlabeled instances using the model
    		\STATE ${\mathbf a}$ = \texttt{Score}(${\mathbf D} \setminus \{x: (x, .) \in {\bf H}_f\}$, $\mathcal{E}$, {\em FSSN})
    		\end{tcolorbox}
    		\STATE // Greedy selection: highest-scoring instance
    		\STATE Let ${\bf q} = {\mathbf x}_i$, where $i = \argmax_{i}(a_i)$
    		\begin{tcolorbox}[colback=green!5!white,colframe=green!75!black]
    		\STATE Get label $y_i \in \{-1,+1\}$ for the highest scoring unlabeled instance ${\mathbf q}$ from human analyst
    		\end{tcolorbox}
    		\STATE // Aggregate set of labeled instances
    		\STATE Set ${\mathbf H}_f = \{({\bf x}_i, y_i)\} \cup {\mathbf H}_f$
    		
    		\STATE
    		\begin{tcolorbox}[colback=green!5!white,colframe=green!75!black]
    		\STATE Update the parameters of {\em FSSN} by minimizing loss in Equation~\ref{eqn:loss_fssn} for aggregate labeled set ${\mathbf H}_f$
    		\end{tcolorbox}
    		\STATE
		\ENDFOR
		\STATE \textbf{return} {\em FSSN, ${\mathbf H}_f$}
	\end{algorithmic}
\end{algorithm}

\begin{align}
	&\ell_{prior}({\bf x})=\sum_{m=1}^M{-b\log(p_m({\bf x})) -(1 - b)\log(1 - p_m({\bf x}))} \label{eqn:prior} \\
	&\ell_A(q; ({\bf x}, y)) = \max(0, y (q - \text{\texttt{Score}$({\bf x})$})) \label{eqn:la_loss} \\
	&\ell_{AAD}({\bf x}, y) = \ell_A(q_{\tau}^{(t-1)}; ({\bf x}, y)) + \ell_A(\text{\texttt{Score}}({\bf x}_{\tau}^{(t-1)}); ({\bf x}, y)) \label{eqn:aad_loss} \\
	&\ell_{FSSN} = \frac{1}{|{\bf H}_f^{(t)}|} \sum_{({\bf x}, y) \in {\bf H}_f^{(t)}} \ell_{AAD}({\bf x}, y) + \frac{\lambda}{|{\bf D}|} \sum_{{\bf x} \in {\bf D}}\ell_{prior}({\bf x})\label{eqn:loss_fssn}
\end{align}

\paragraph{Feature Space Suppression Network (FSSN).} The FSSN is a neural network with $M$ sigmoid activation nodes in its output layer, where each output node is paired with an ensemble member. It takes as input an instance from the original feature space and outputs the relevance of each detector for that instance. We denote the relevance of the $m^{th}$ detector to instance ${\bf x}$ by $p_m({\bf x})$. The FSSN is primed using the cross-entropy loss in Equation~\ref{eqn:prior} such that it outputs the same probability $b \in (0, 1)$ at all the output nodes for each data instance in ${\bf D}$. This loss acts as a prior on the relevance of detectors in ensemble. When all detectors have the same relevance, the final anomaly score simply corresponds to the average score across all detectors (up to a multiplicative constant), and is \textbf{a good starting point for active learning.}

After FSSN is primed, it automatically learns the relevance of the detectors based on label feedback from human analyst using the combined loss $\ell_{FSSN}$ in Equation~\ref{eqn:loss_fssn}, where $\lambda$ is the trade-off parameter. We set the value of $\lambda$ to 1 in all our experiments. $\mathbf{H}_f^{(t)} \subseteq {\bf D}$ in Equation~\ref{eqn:loss_fssn} denotes the total set of instances labeled by the analyst after $t$ feedback iterations. ${\mathbf x}_{\tau}^{(t-1)}$ and $q_{\tau}^{(t-1)}$ denote the instance ranked at the $\tau$-th quantile and its score after the $(t-1)$-th feedback iteration. $\ell_{A}$ encourages the scores of anomalies in $\mathbf{H}_f$ to be higher than that of $q$, and the scores of nominals in $\mathbf{H}_f$ to be lower.



\section{Explanations for Anomalies with GLAD}

To help the analyst understand the results of active anomaly detection system, we now introduce the concept of ``explanations'' in the context of GLAD model. 

\begin{itemize}
	\item \textbf{Explanation:} An explanation outputs a reason why a specific data instance was assigned a high anomaly score. Generally, we limit the scope of an explanation to one data instance. The main application is to diagnose the model: whether the anomaly detector(s) are working as expected or not.
\end{itemize}

\begin{figure}
	\centering
	\includegraphics[width=0.23\textwidth]{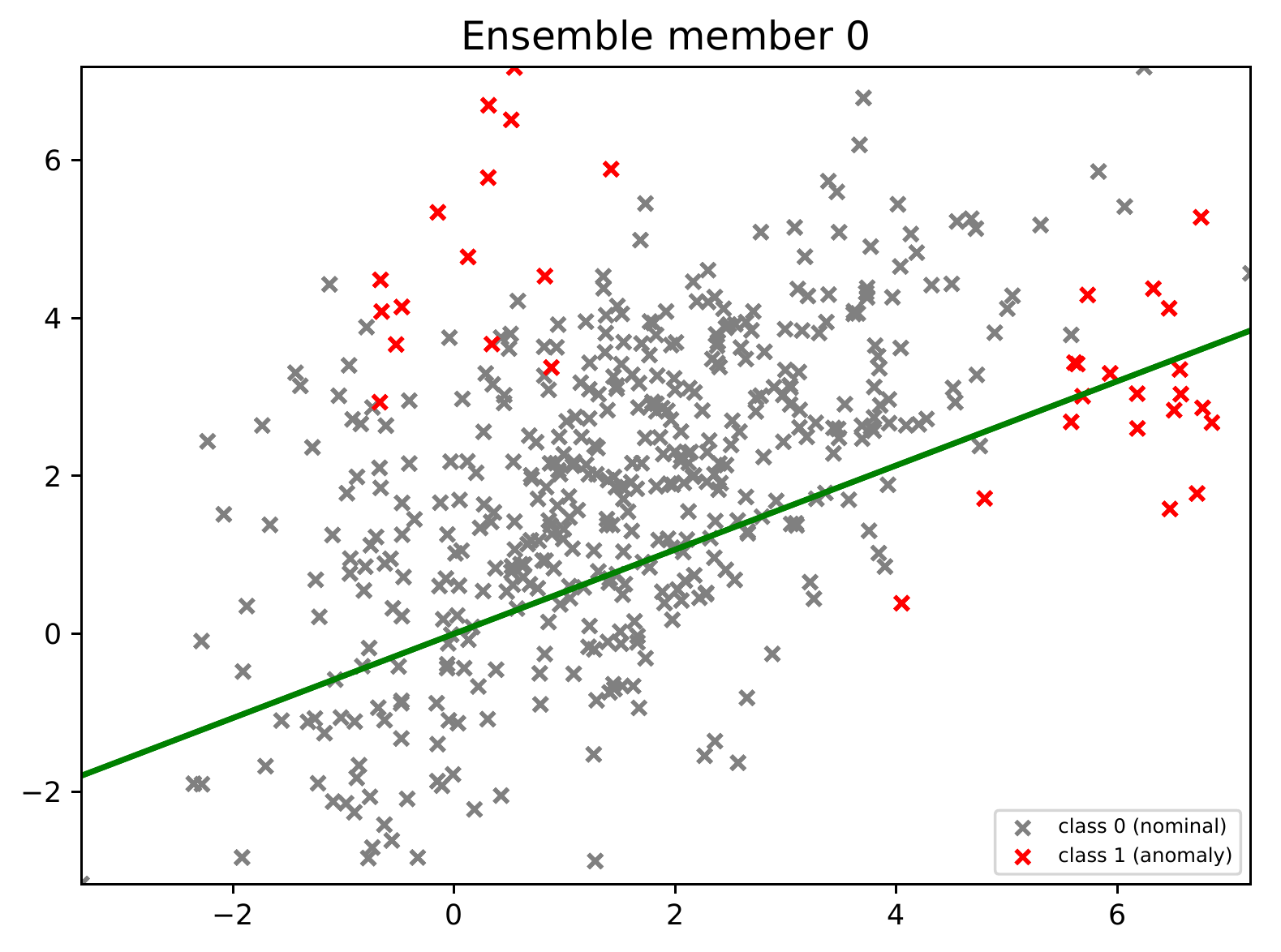}
	\includegraphics[width=0.23\textwidth]{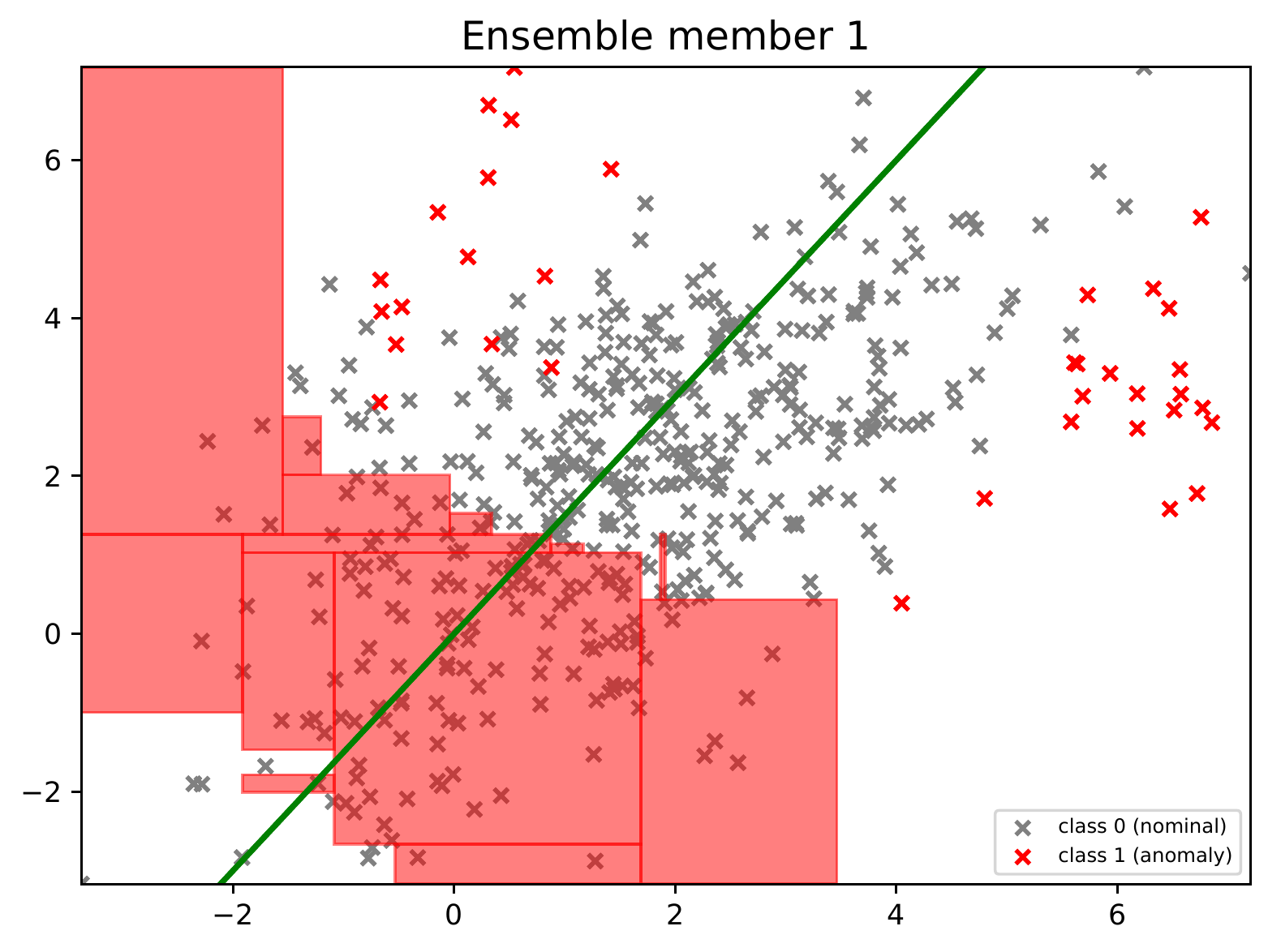}\\
	\includegraphics[width=0.23\textwidth]{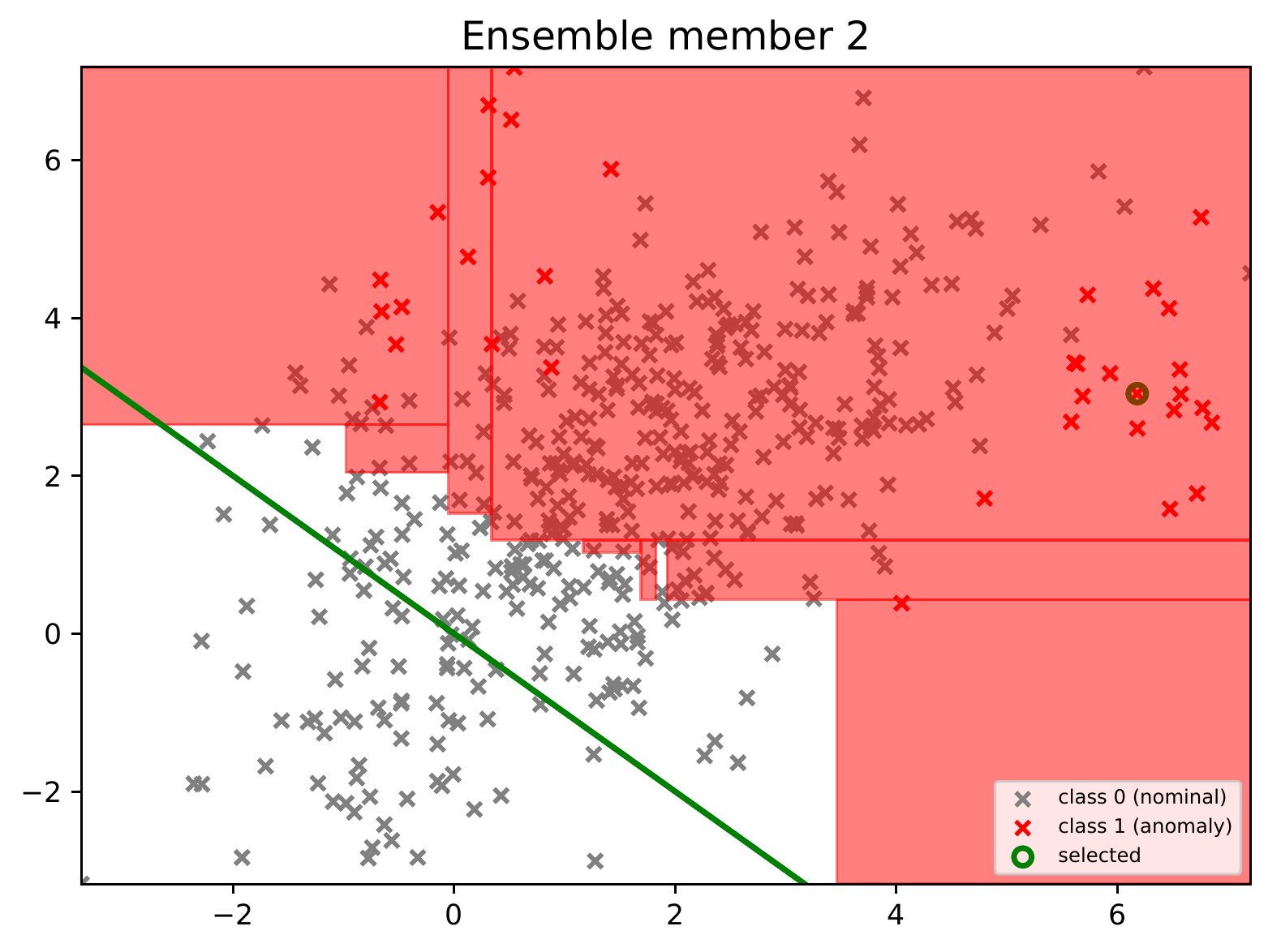}
	\includegraphics[width=0.23\textwidth]{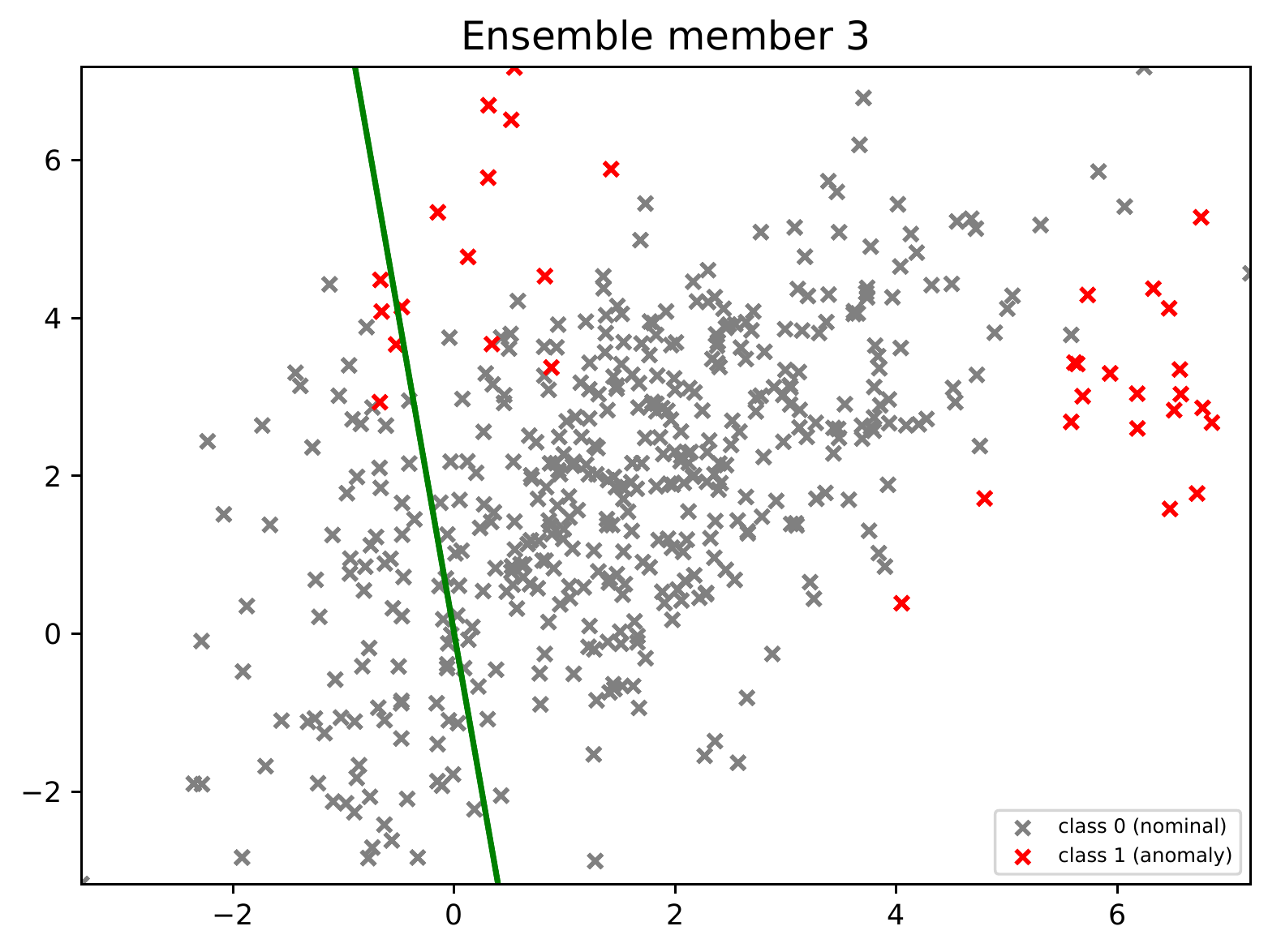}
	\caption{Most relevant ensemble members in subspaces inferred with GLAD after 30 feedback iterations. There are four members (i.e., LODA projections) in our current example. 
	Note that members 1 and 2 were found relevant in subspaces which have mostly nominal instances. This is because they correctly assigned low anomaly scores to instances in those subspaces. The last member (member 3) did not rank as the top-most relevant detector for any instance; hence, it does not have any region marked in red. The point circled in green (in bottom left plot) is a test instance. Ensemble member 2 was found to be the most relevant for this instance.}
	\label{fig:glad_relevance}
\end{figure}

GLAD assumes that the anomaly detectors in the ensemble can be arbitrary (homogeneous or heterogeneous). The best it can offer as an explanation is to output the member which is most relevant for a test data instance. With this in mind, we can employ the following approach to generate explanations:

\begin{enumerate}

	\item Employ the FSSN network to predict the relevance of individual ensemble members on the complete dataset. It is important to note that the relevance of a detector is different from the anomaly score(s) it assigns. A detector which is relevant in a particular subspace predicts the labels of instances in that subspace correctly irrespective of whether those instances are anomalies or nominals.
	
	\item Find the instances for each ensemble member for which that detector is the most relevant. Mark these instances as positive and the rest as negative.
	
	\item Train a separate decision tree for each member to separate the corresponding positives and negatives. This describes the subspaces where each ensemble member is relevant. Figure~\ref{fig:glad_relevance} illustrates this idea on the {\em Toy} dataset.
	
	\item When asked to explain the anomaly score for a given test instance:
	
	\begin{enumerate}
		\item Use FSSN network to identify the most relevant ensemble member for the test instance.
		\item Employ a model agnostic explanation technique such as LIME \cite{ribeiro:2016} or ANCHOR \cite{ribeiro:2018} to generate the explanation using the most relevant ensemble member. As a simple illustration, we trained GLAD on the synthetic dataset and a LODA ensemble with four projections. After $30$ feedback iterations, the unlabeled instance at $(6.12, 3.04)$ had the highest anomaly score. We used LIME to explain its anomaly score. LIME explanation is shown below:
		\begin{verbatim}
		('2.16 < y <= 3.31', -0.4253)
		('x > 2.65', 0.3406)
		\end{verbatim}
		Here the explanation \texttt{2.16 < y <= 3.31} from member $2$ has the highest absolute weight and hence, explains most of the anomaly score.
	\end{enumerate}
\end{enumerate}

Since most aspects of explanations are qualitative, we leave their evaluation on real-world data to future work.

\section{Experiments and Results}
\label{sec:experiments}

In this section, we describe our experimental setup and present results on both synthetic and real-world datasets.

\paragraph{LODA based Anomaly Detector.} For our anomaly detector, we employ the \textit{LODA} algorithm \cite{pevny:2015}, which is an ensemble  ${\bf \mathcal{E}} = \{\mathcal{D}_m\}_{m=1}^M$ of $M$ one-dimensional histogram density estimators computed from sparse random projections. Each projection $\mathcal{D}_m$ is defined by a sparse $d$-dimensional random vector $\bf{\beta}_m$. LODA projects each data point onto the real line according to $\bf{\beta}_m^{\top}{\bf x}$ and then forms a histogram density estimator $f_m$. The anomaly score assigned to a given instance ${\bf x}$ is the mean negative log density: \texttt{Score}(${\bf x}$) = $\frac{1}{M} \sum_{m=1}^M{s_m({\bf x})}$, where, $s_m({\bf x}) \triangleq -\log(f_m({\bf x}))$.

LODA gives equal weights to all projections. Since the projections are selected at random, there is no guarantee that every projection is good at isolating anomalies uniformly across the entire input feature space. LODA-AAD \cite{das:2016} was proposed to integrate label feedback from a human analyst by learning a better weight vector ${\bf w}$ that assigns weights proportional to the usefulness of the projections. In this case, the learned weights are \textit{global}, i.e., they are fixed across the entire input feature space. In contrast, we employ GLAD to learn the \textit{local} relevance of each detector in the input space using the label feedback.
\begin{figure}[]
	\centering
	\includegraphics[width=0.48\textwidth]{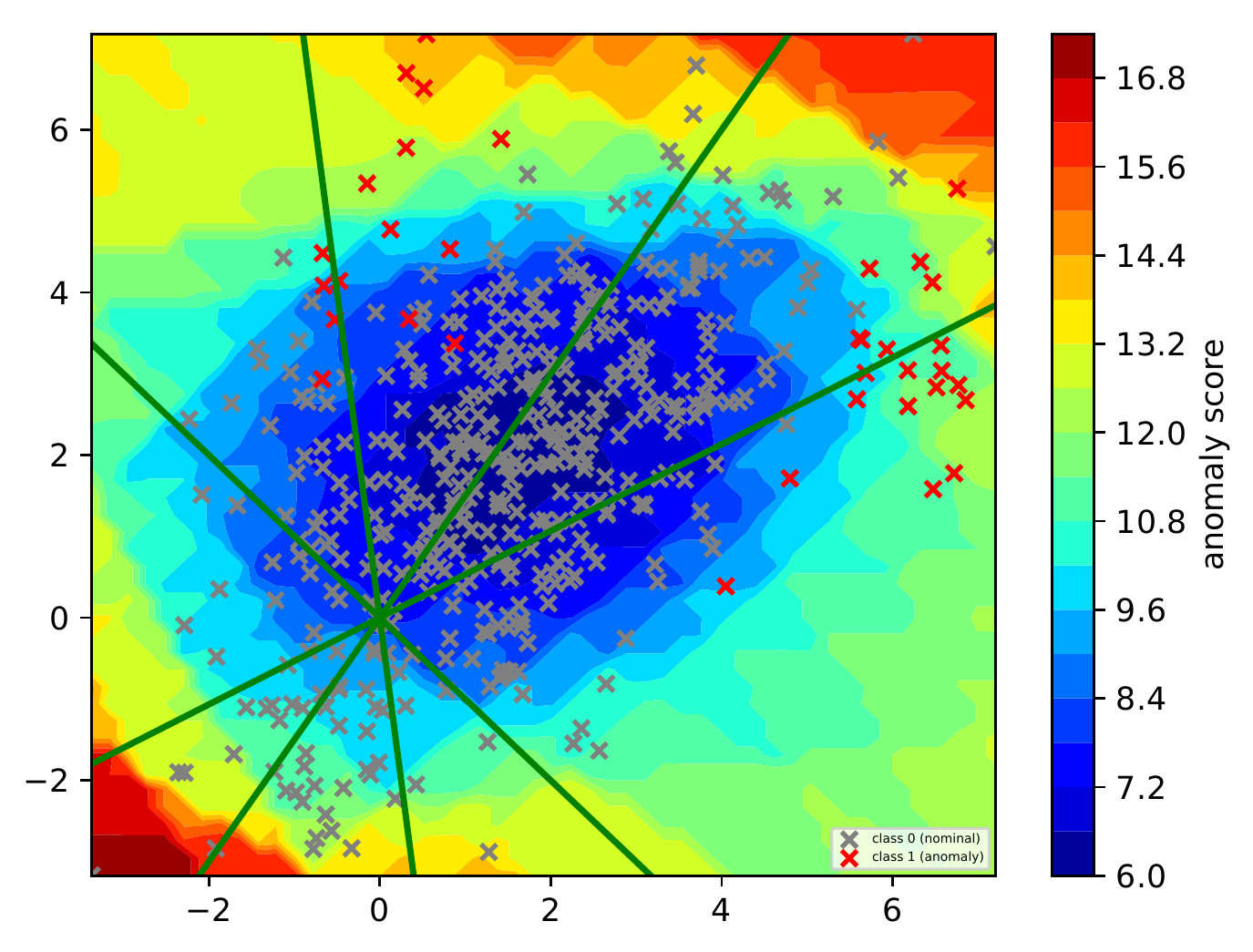}
	\caption{Baseline LODA score contours on \textit{toy dataset.} Here we presented LODA with four projections (green lines) applied to the same dataset. Red represents locations with higher anomaly score and blue for nominals.}
	\label{fig:loda_all}
\end{figure}

\begin{figure}[]
	\centering
	\includegraphics[width=0.48\textwidth]{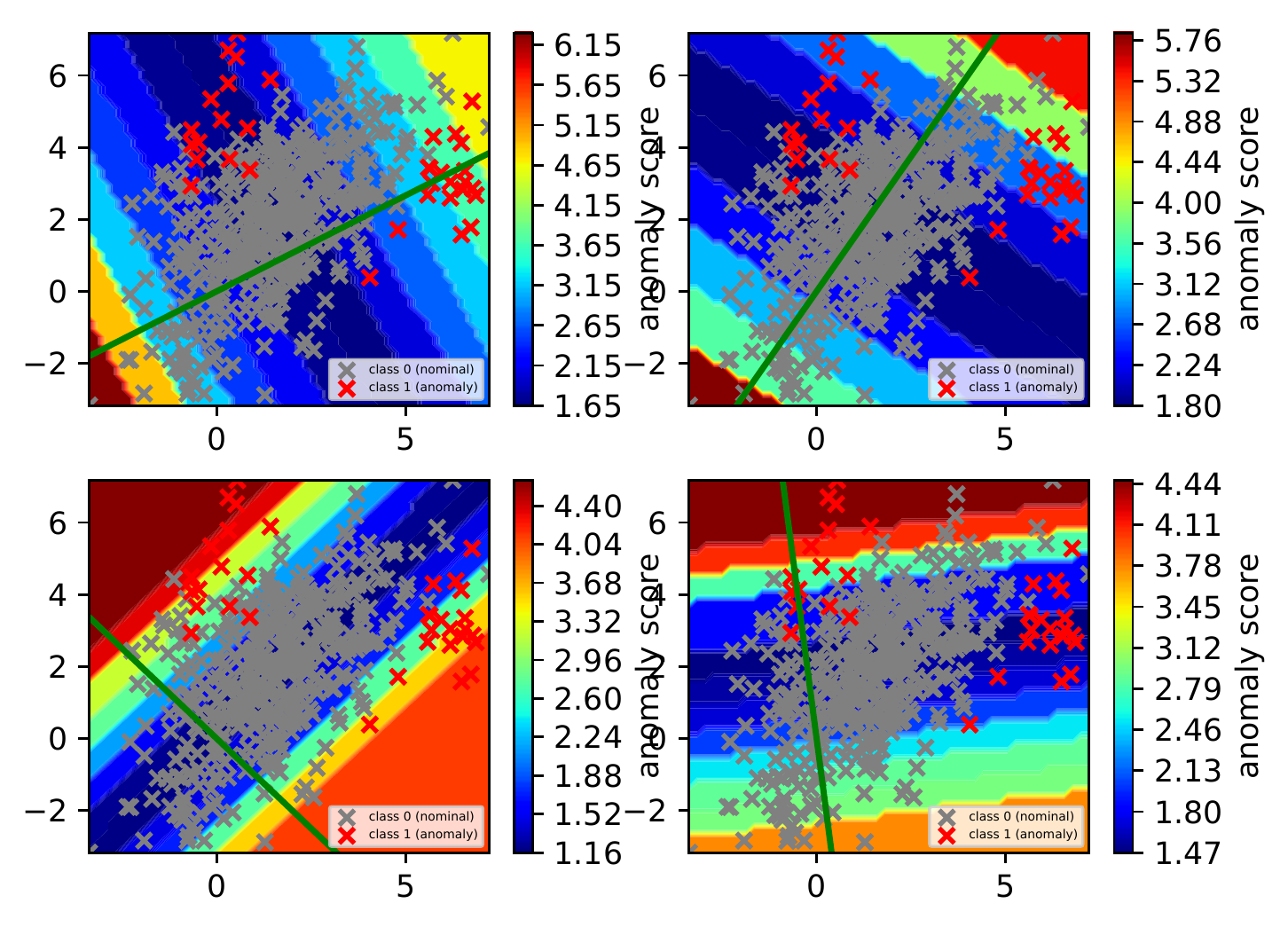}
	\caption{Score contours for each \textit{LODA} projection. More red zones are possible places where anomalous candidates could be present. And dark blue zones are representing less possible places for anomalous data. For illustration purpose we are presenting 4 different projection.}
	\label{fig:loda_scores}
\end{figure}

\begin{figure}[]
	\centering
	\includegraphics[width=0.48\textwidth]{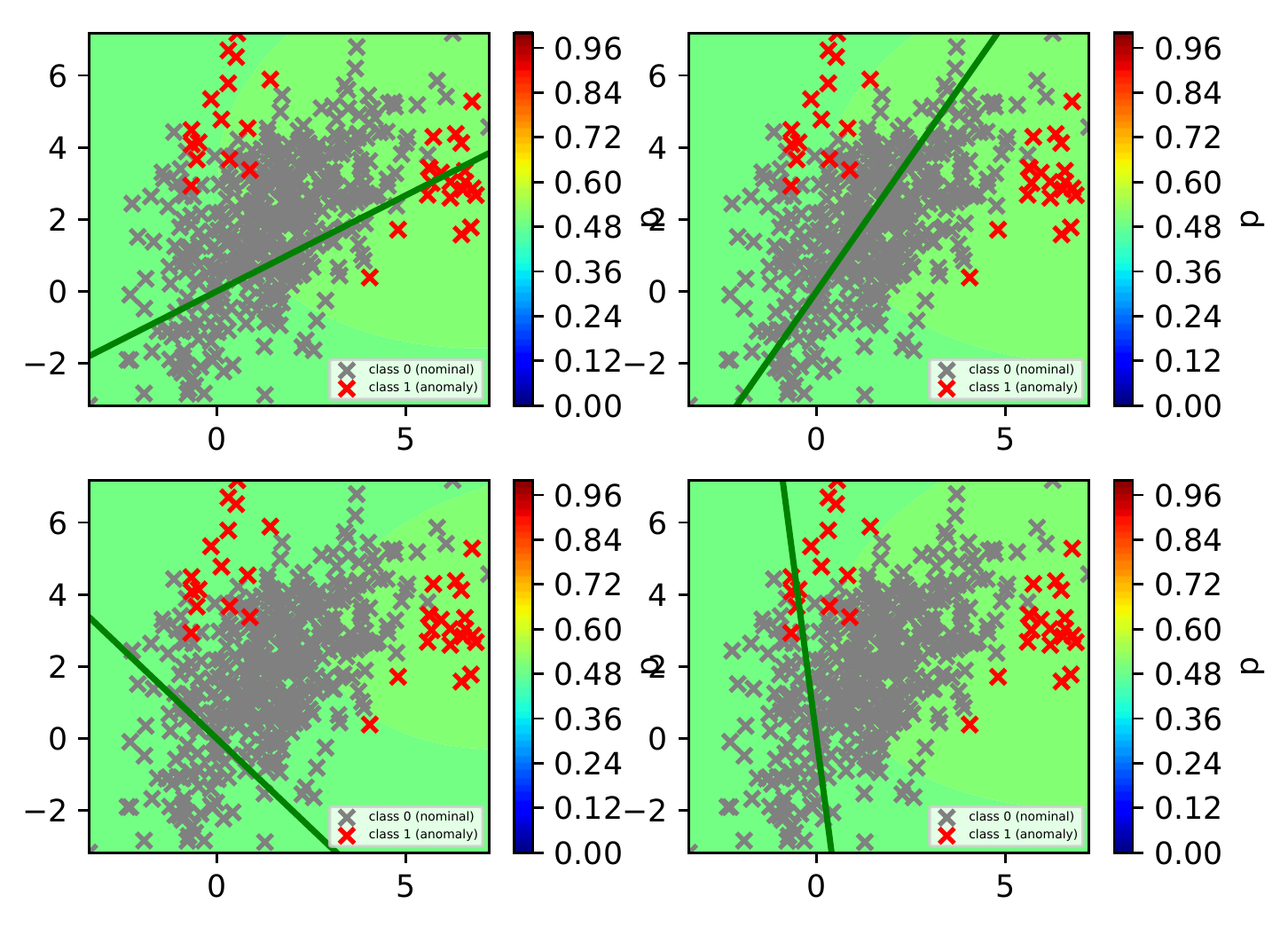}
	\caption{The red `$\times$' are true anomalies and grey `$\times$' are true nominals. The output nodes of the FSSN are initially primed to return a relevance of $0.5$ everywhere in the feature space.}
	\label{fig:loda_uniform}
\end{figure}

\paragraph{FSSN Details.} We introduced a shallow neural network with $\max(50, 3M)$ hidden nodes for all our test datasets, where $M$ is the number of ensemble members (i.e., LODA projections). The network is retrained after receiving each label feedback. This retraining cycles over the entire dataset (labeled and unlabeled) once. Since the labeled instances are very few, we up-sample the labeled data five times. We also employ $L_2$-regularization for training the weights of the neural network.
\begin{figure}[]
	\centering
	\subfigure[Relevance after 30 feedback iterations]{\includegraphics[width=0.48\textwidth]{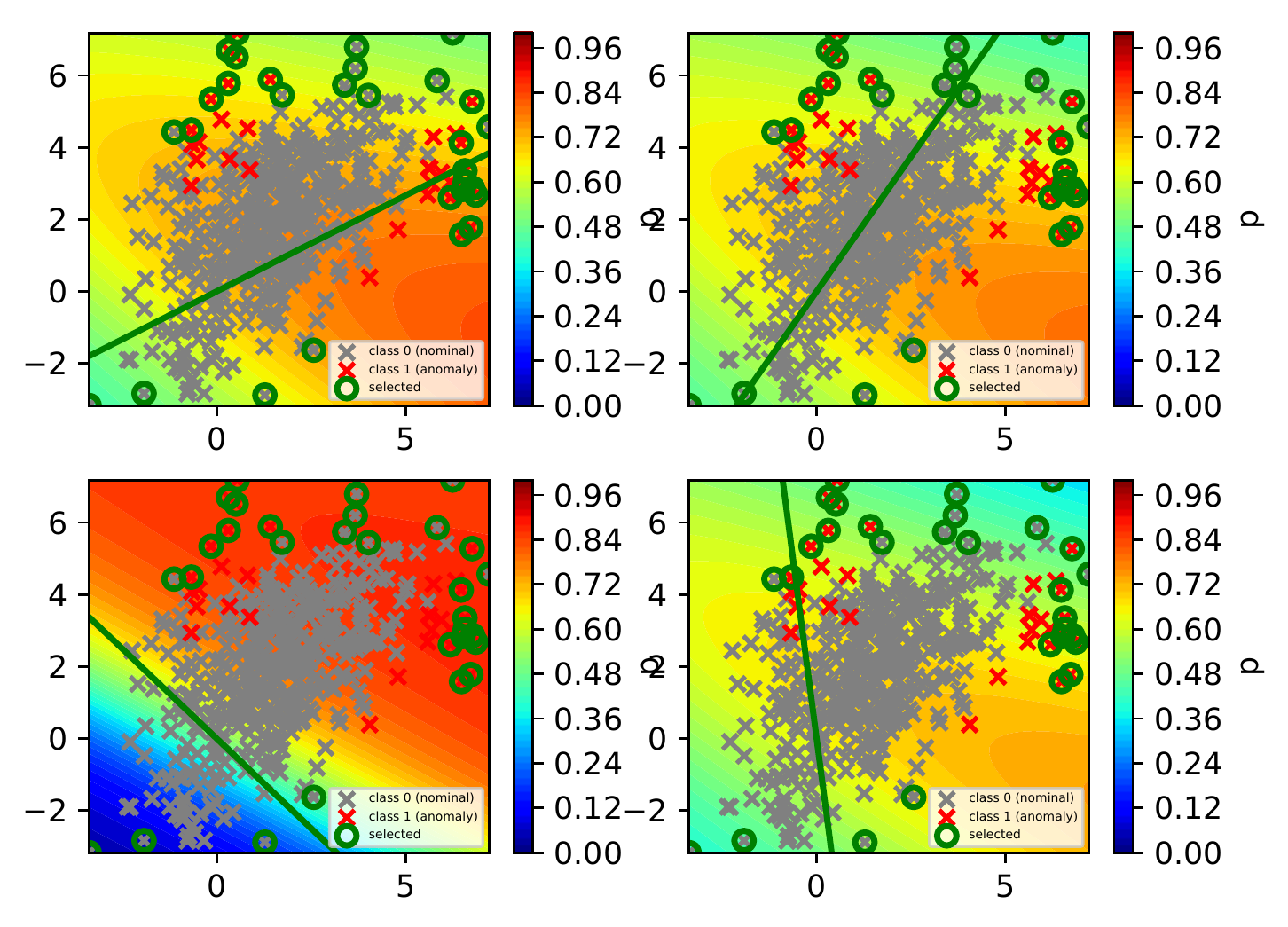} \label{fig:loda_relevance}}
	\caption{The points circled in green were shown to the analyst for labeling, one per feedback iteration. After $30$ feedback iterations, the bottom left projection was found to be most relevant in the top-right half-space, whereas it is completely irrelevant in the bottom-left half-space. Other projections were less relevant in most parts of the feature space.}
	\label{fig:dataset}
\end{figure}
\begin{table}[]
\centering
\begin{tabular}{|c|r|r|r|}
\hline
\textbf{Dataset} & \textbf{Total} & \textbf{Dims} & \textbf{\# Anomalies(\%)} \\ \hline
Abalone          & 1920           & 9             & 29 (1.5\%)                \\ \hline
ANN-Thyroid-1v3  & 3251           & 21            & 73 (2.25\%)               \\ \hline
Cardiotocography & 1700           & 22            & 45 (2.65\%)               \\ \hline
KDD-Cup-99       & 63009          & 91            & 2416 (3.83\%)             \\ \hline
Mammography      & 11183          & 6             & 260 (2.32\%)              \\ \hline
Yeast            & 1191           & 8             & 55 (4.6\%)                \\ \hline
\end{tabular}
\caption{Description of benchmark datasets.}
\label{tab:real-dataset}
\end{table}

\paragraph{Synthetic Experiments.} The \textit{Toy} dataset and the corresponding LODA ensembles have been shown in Figure~\ref{fig:loda_all}. Figure~\ref{fig:loda_scores} illustrates the aspect that detectors are varying in quality. Figure~\ref{fig:loda_relevance} shows that GLAD learns useful relevance information that can be of help to the analyst.

\begin{figure}[htb]
	\centering 
	\subfigure[Abalone]{
		\includegraphics[width=0.22\textwidth]{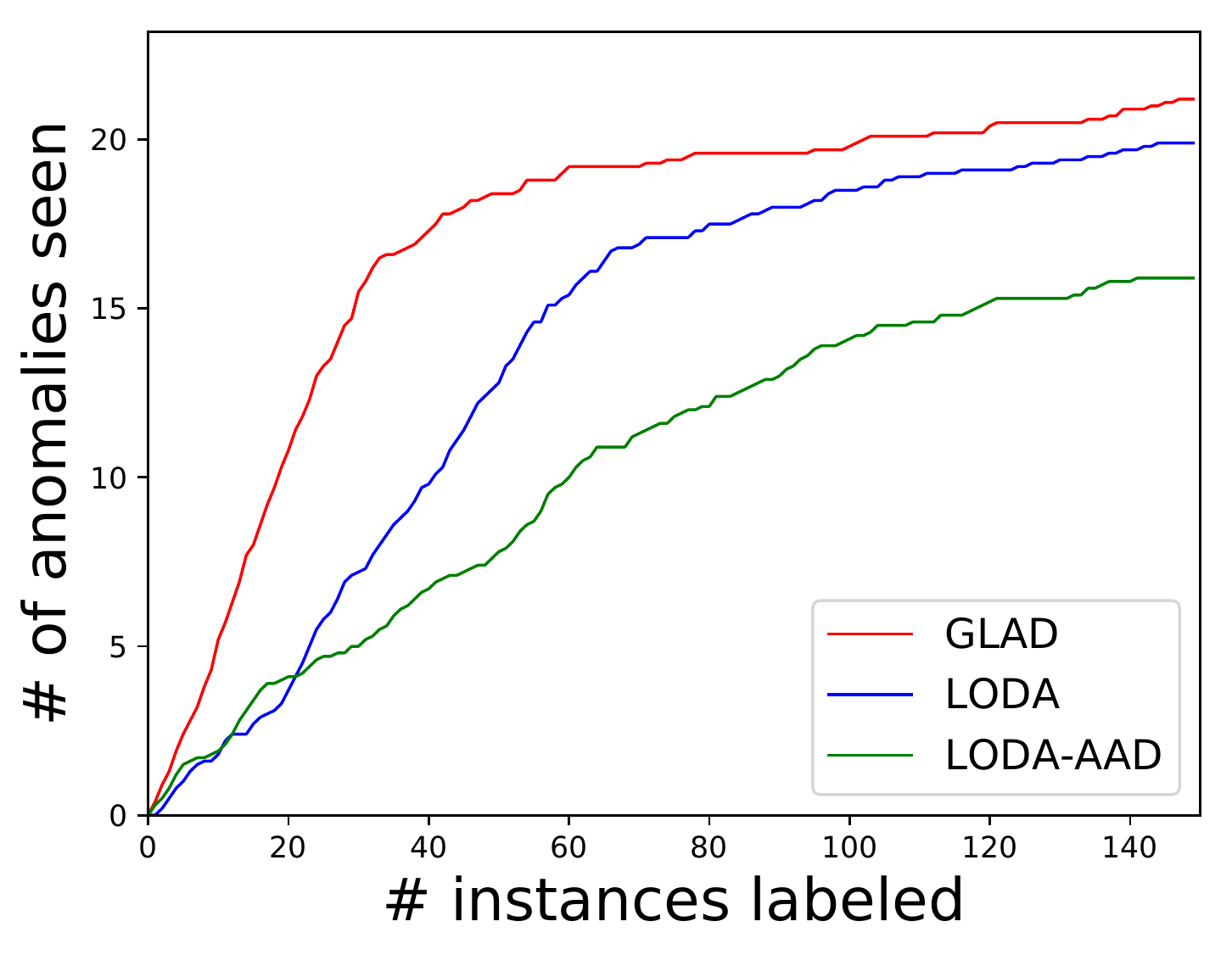}
		\label{fig:afss_abalone}}
	\subfigure[ANN-Thyroid-1v3]{
		\includegraphics[width=0.22\textwidth]{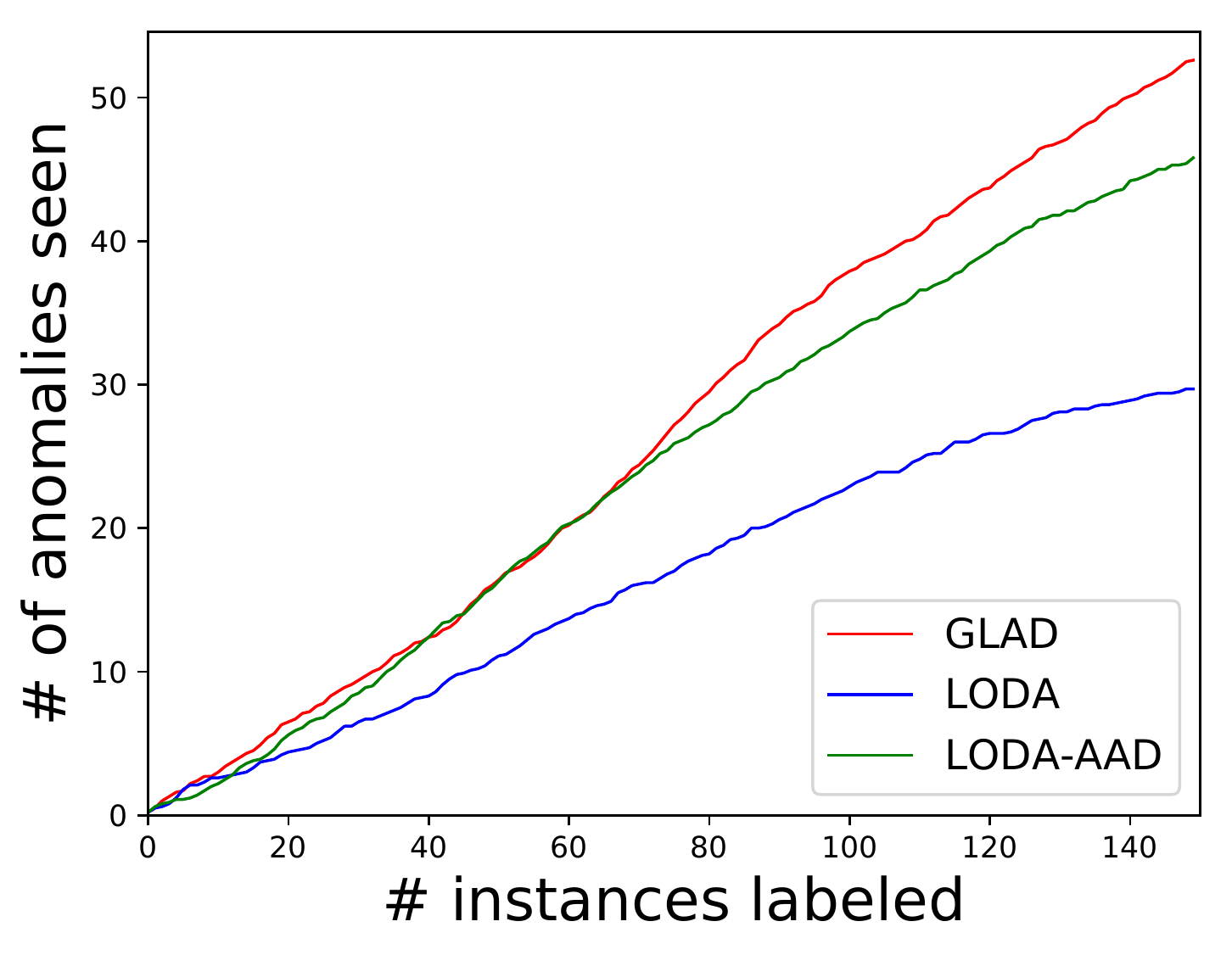}
		\label{fig:afss_ann_thyroid_1v3}}\\
	\subfigure[Cardiotocography]{
		\includegraphics[width=0.22\textwidth]{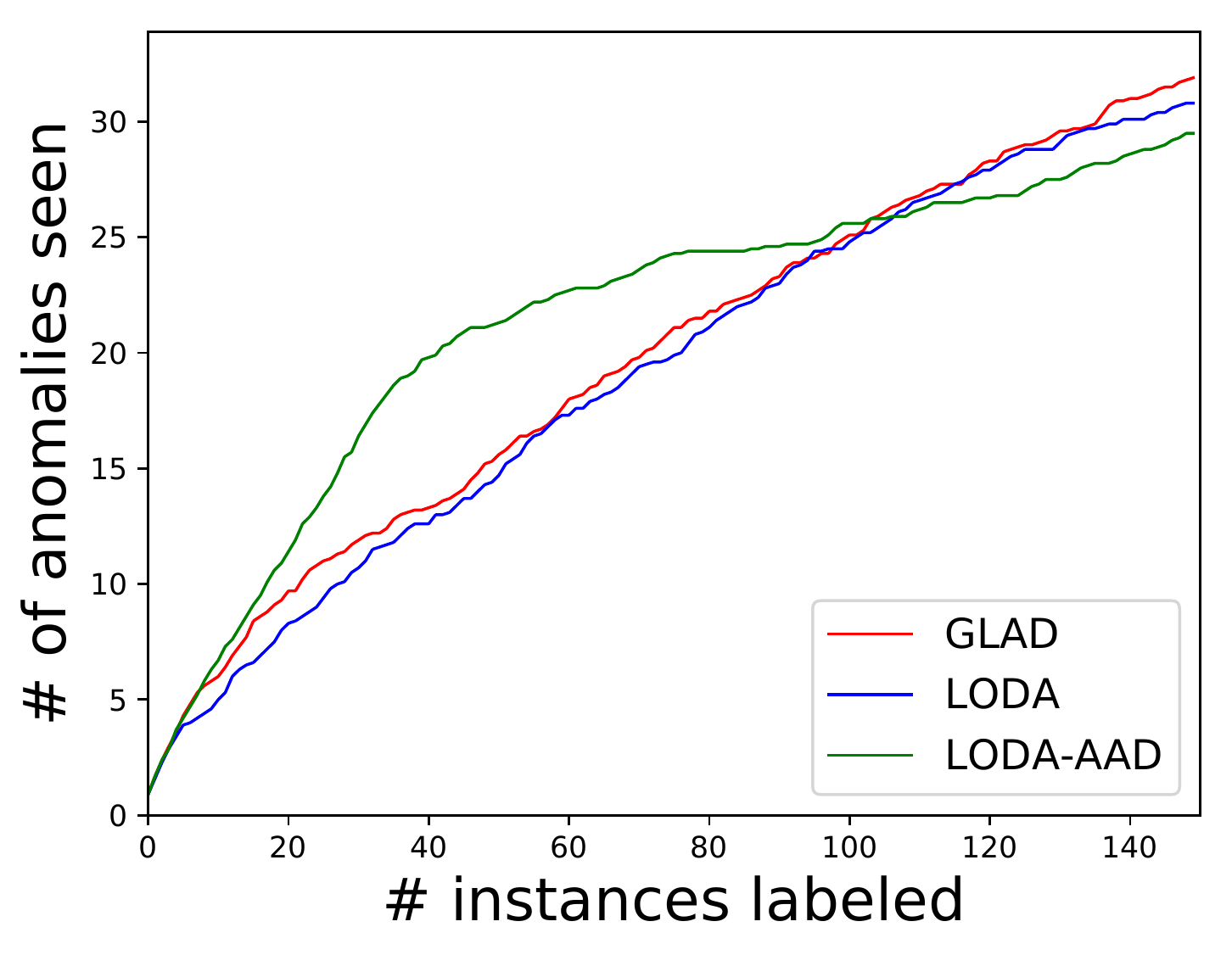}
		\label{fig:afss_cardiotocography_1}}
	\subfigure[Yeast]{
		\includegraphics[width=0.22\textwidth]{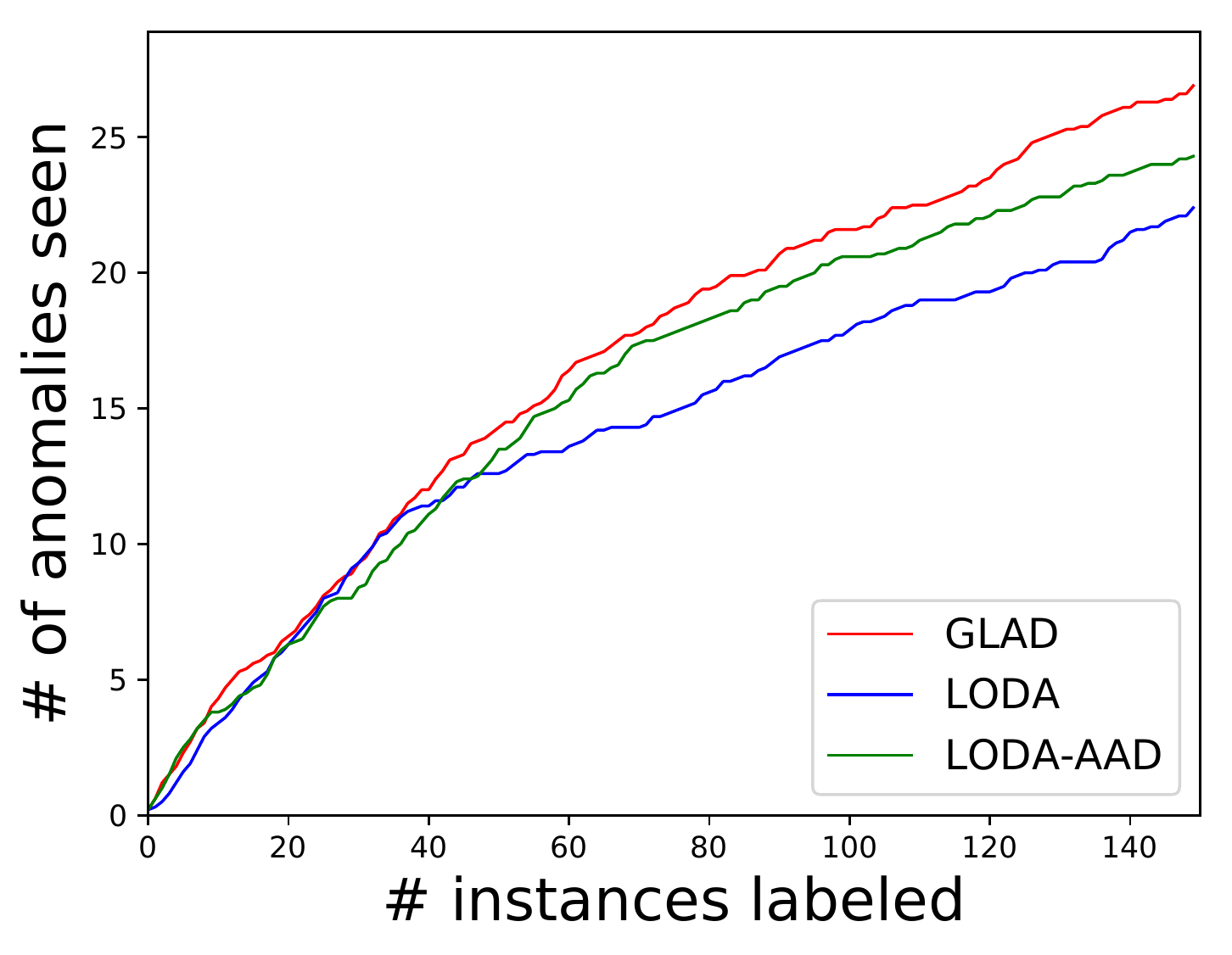}
		\label{fig:afss_yeast}}\\
	\subfigure[Mammography]{
		\includegraphics[width=0.22\textwidth]{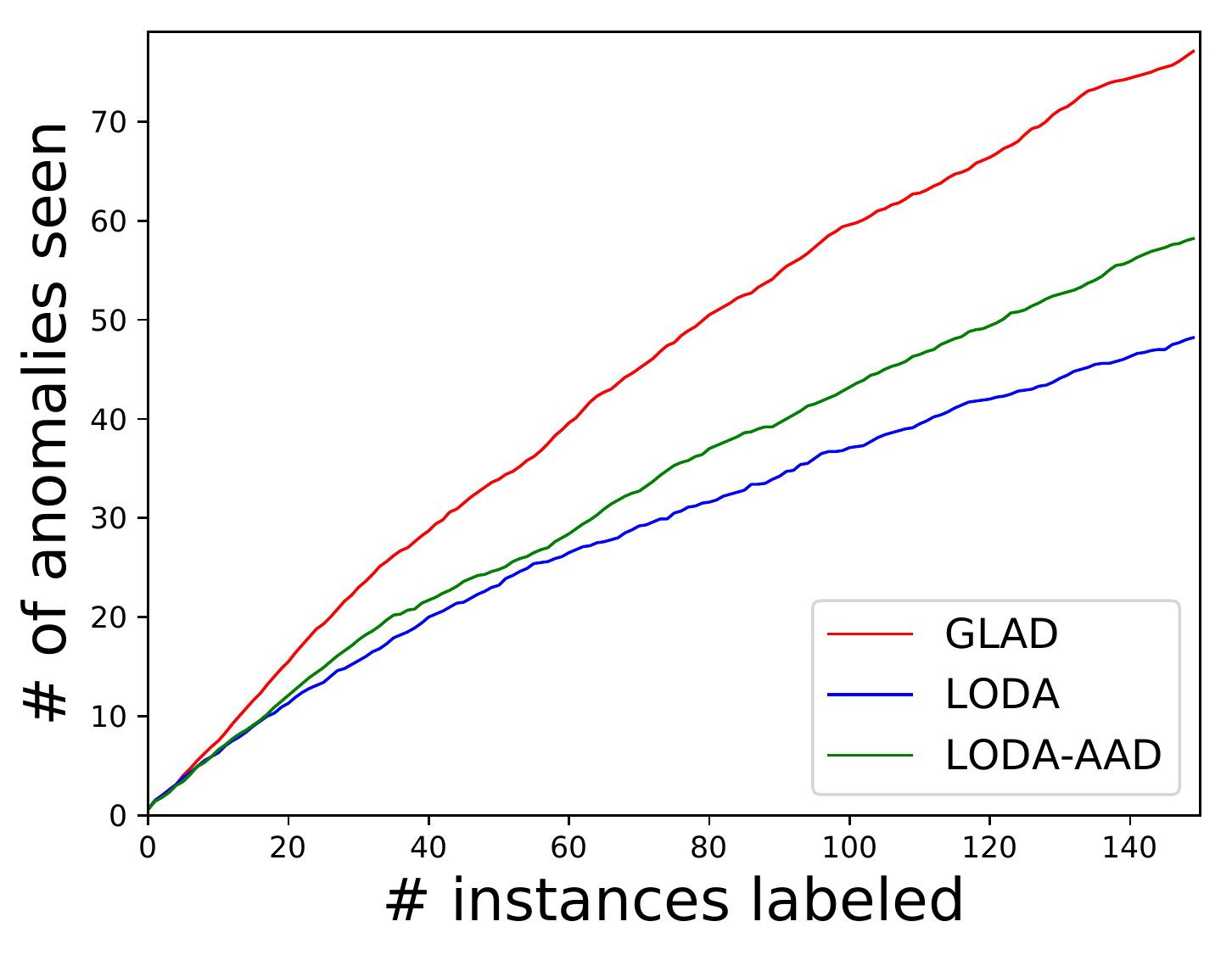}
		\label{fig:afss_mammography}}
	\subfigure[KDD-Cup-99]{
		\includegraphics[width=0.22\textwidth]{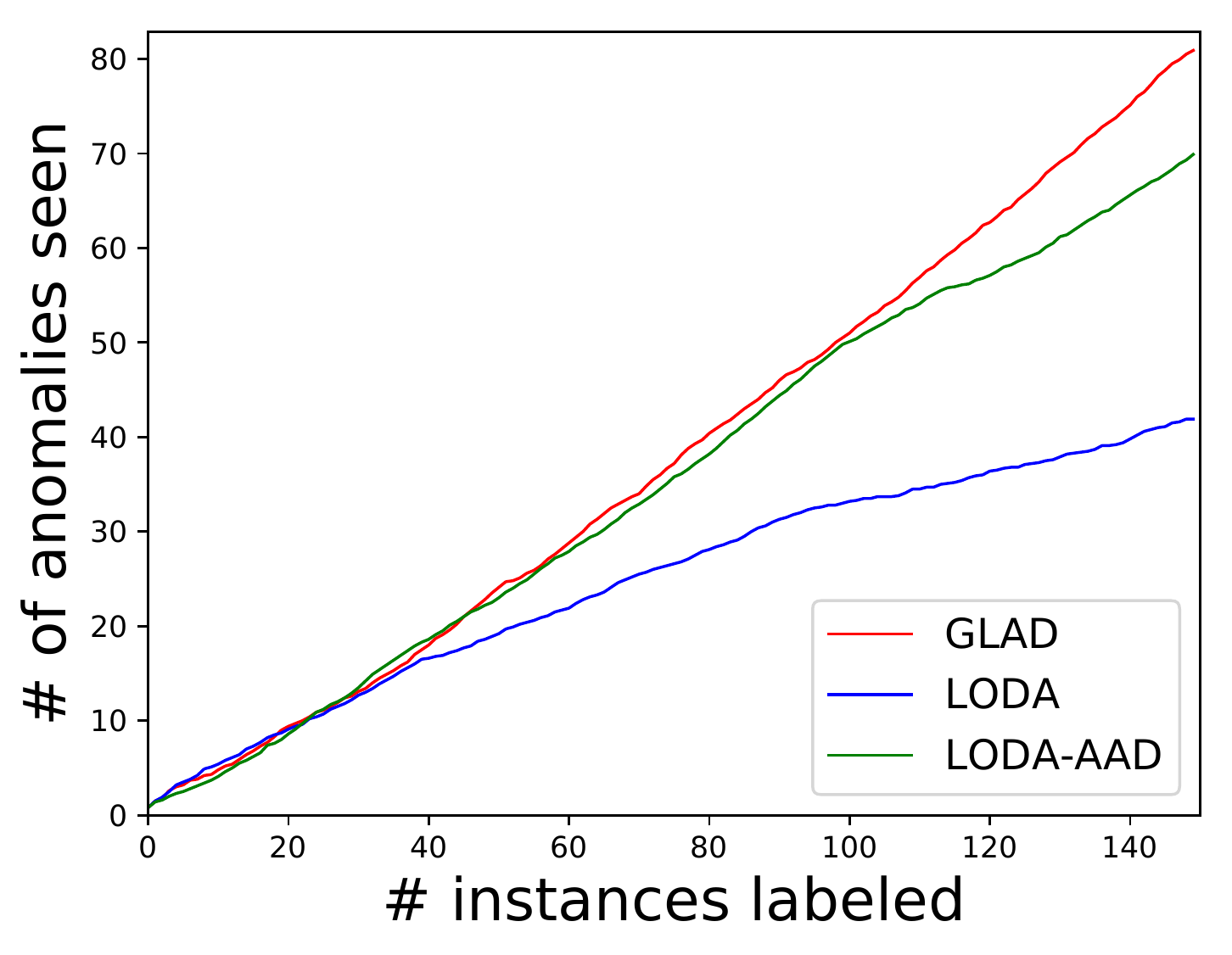}
		\label{fig:afss_kddcup}}\\
	\caption{Results on real-world datasets. Number of anomalies discovered as a function of the number of labeled queries. Results were averaged over $10$ different runs. The higher the curve the better the performance.}
	\label{fig:afss_all}
\end{figure}

\paragraph{Real-world Experiments.} We demonstrate the effectiveness of GLAD on most of the datasets from Table \ref{tab:real-dataset} used in \cite{das:2018}. Since GLAD is most relevant when the anomaly detectors are specialized and fewer in number, we employ a LODA ensemble with maximum $15$ projections. In Figure~\ref{fig:afss_all}, we observe that GLAD outperforms both the baseline LODA as well as LODA-AAD which weights the ensemble members globally.

\section{Discussion}

It is well-known that there exists no universally applicable anomaly detector. However, sometimes a few easy-to-understand detectors meet most needs of users. Therefore, they should not be marginalized just because they fail in some special cases. Our proposed approach \textit{GLAD} learns when the detectors are relevant. Therefore makes it more likely that the preferred detectors of users will be applied or suppressed as needed. Finally, it can provide explanations for the end user to acquire better understanding about the anomalies.
\section{Acknowledgements}
This work was supported in part by contract W911NF15-1-0461 with the US Defense Advanced Research Projects Agency (DARPA) Communicating with Computers Program and the Army Research Office (ARO), and the National Science Foundation (NSF) grant IIS-1543656 . The views expressed are those of the authors and do not reflect the official policy or position of the Department of Defense or NSF or the U.S. Government.  
\bibliography{glocalized}

\begin{thebibliography}{32}
\providecommand{\natexlab}[1]{#1}
\providecommand{\url}[1]{\texttt{#1}}
\expandafter\ifx\csname urlstyle\endcsname\relax
  \providecommand{\doi}[1]{doi: #1}\else
  \providecommand{\doi}{doi: \begingroup \urlstyle{rm}\Url}\fi

\bibitem[Abe et~al.(2006)Abe, Zadrozny, and Langford]{abe:2006}
Abe, N., Zadrozny, B., and Langford, J.
\newblock Outlier detection by active learning.
\newblock In \emph{Proceedings of the Twelth {ACM} {SIGKDD} International
  Conference on Knowledge Discovery and Data Mining {(KDD)}}, pp.\  504--509,
  2006.

\bibitem[Aggarwal \& Sathe(2017)Aggarwal and Sathe]{aggarwal:2017}
Aggarwal, C.~C. and Sathe, S.
\newblock \emph{Outlier Ensembles}.
\newblock Springer, 2017.

\bibitem[Almgren \& Jonsson(2004)Almgren and Jonsson]{almgren:2004}
Almgren, M. and Jonsson, E.
\newblock Using active learning in intrusion detection.
\newblock In \emph{17th {IEEE} Computer Security Foundations Workshop,
  {(CSFW)}}, pp.\ ~88, 2004.

\bibitem[Blanchard et~al.(2010)Blanchard, Lee, and Scott]{blanchard2010semi}
Blanchard, G., Lee, G., and Scott, C.
\newblock Semi-supervised novelty detection.
\newblock \emph{Journal of Machine Learning Research}, 11\penalty0
  (Nov):\penalty0 2973--3009, 2010.

\bibitem[Breunig et~al.(2000)Breunig, Kriegel, Ng, and Sander]{breunig:00}
Breunig, M.~M., Kriegel, H.-P., Ng, R.~T., and Sander, J.
\newblock Lof: Identifying density-based local outliers.
\newblock In \emph{ACM SIGMOD International Conference on Management of Data},
  2000.

\bibitem[Chapelle et~al.(2009)Chapelle, Scholkopf, and Zien]{chapelle2009semi}
Chapelle, O., Scholkopf, B., and Zien, A.
\newblock Semi-supervised learning (chapelle, o. et al., eds.; 2006)[book
  reviews].
\newblock \emph{IEEE Transactions on Neural Networks}, 20\penalty0
  (3):\penalty0 542--542, 2009.

\bibitem[Das et~al.(2016)Das, Wong, Dietterich, Fern, and Emmott]{das:2016}
Das, S., Wong, W.-K., Dietterich, T.~G., Fern, A., and Emmott, A.
\newblock Incorporating expert feedback into active anomaly discovery.
\newblock In \emph{IEEE ICDM}, 2016.

\bibitem[Das et~al.(2017)Das, Wong, Fern, Dietterich, and Siddiqui]{das:2017}
Das, S., Wong, W.-K., Fern, A., Dietterich, T.~G., and Siddiqui, M.~A.
\newblock Incorporating expert feedback into tree-based anomaly detection.
\newblock In \emph{KDD IDEA Workshop}, 2017.

\bibitem[Das et~al.(2018)Das, Islam, Jayakodi, and Doppa]{das:2018}
Das, S., Islam, M.~R., Jayakodi, N.~K., and Doppa, J.~R.
\newblock Active anomaly detection via ensembles.
\newblock \emph{arXiv:1809.06477}, 2018.
\newblock [Online; accessed 19-Sep-2018].

\bibitem[Das et~al.(2020)Das, Wong, Dietterich, Fern, and Emmott]{TKDD-2020}
Das, S., Wong, W.-K., Dietterich, T., Fern, A., and Emmott, A.
\newblock Discovering anomalies by incorporating feedback from an expert.
\newblock \emph{ACM Trans. Knowl. Discov. Data}, 14\penalty0 (4), June 2020.
\newblock ISSN 1556-4681.
\newblock \doi{10.1145/3396608}.
\newblock URL \url{https://doi.org/10.1145/3396608}.

\bibitem[Doshi-Velez \& Kim(2017)Doshi-Velez and Kim]{doshi-velez:2017}
Doshi-Velez, F. and Kim, B.
\newblock Towards a rigorous science of interpretable machine learning.
\newblock \emph{arXiv:1702.08608}, 2017.

\bibitem[Emmott et~al.(2015)Emmott, Das, Dietterich, Fern, and
  Wong]{emmott:2015}
Emmott, A., Das, S., Dietterich, T.~G., Fern, A., and Wong, W.
\newblock Systematic construction of anomaly detection benchmarks from real
  data.
\newblock \emph{CoRR}, abs/1503.01158, 2015.
\newblock URL \url{http://arxiv.org/abs/1503.01158}.

\bibitem[F{\"{u}}rnkranz et~al.(2012)F{\"{u}}rnkranz, Gamberger, and
  Lavrac]{frnkranz:2012}
F{\"{u}}rnkranz, J., Gamberger, D., and Lavrac, N.
\newblock \emph{Foundations of Rule Learning}.
\newblock Cognitive Technologies. Springer, 2012.

\bibitem[Goh \& Rudin(2014)Goh and Rudin]{goh:2014}
Goh, S.~T. and Rudin, C.
\newblock Box drawings for learning with imbalanced data.
\newblock In \emph{The 20th {ACM} {SIGKDD} International Conference on
  Knowledge Discovery and Data Mining, {(KDD)}}, pp.\  333--342, 2014.

\bibitem[G{\"o}rnitz et~al.(2013)G{\"o}rnitz, Kloft, Rieck, and
  Brefeld]{Grnitz2013TowardSA}
G{\"o}rnitz, N., Kloft, M., Rieck, K., and Brefeld, U.
\newblock Toward supervised anomaly detection.
\newblock \emph{J. Artif. Intell. Res.}, 46:\penalty0 235--262, 2013.

\bibitem[Guha et~al.(2016)Guha, Mishra, Roy, and Schrijvers]{guha:2016}
Guha, S., Mishra, N., Roy, G., and Schrijvers, O.
\newblock Robust random cut forest based anomaly detection on streams.
\newblock In \emph{ICML}, 2016.

\bibitem[He \& Carbonell(2008)He and Carbonell]{he:2008}
He, J. and Carbonell, J.~G.
\newblock Nearest-neighbor-based active learning for rare category detection.
\newblock In \emph{NIPS}, 2008.

\bibitem[Jimenez(1998)]{jimenez:1998}
Jimenez, D.
\newblock Dynamically weighted ensemble neural networks for classification.
\newblock In \emph{In Proceedings of the 1998 International Joint Conference on
  Neural Networks}, pp.\  753--756, 1998.

\bibitem[Letham et~al.(2015)Letham, Rudin, McCormick, and Madigan]{letham:2015}
Letham, B., Rudin, C., McCormick, T.~H., and Madigan, D.
\newblock Interpretable classifiers using rules and bayesian analysis: Building
  a better stroke prediction model, 2015.
\newblock URL \url{http://arxiv.org/abs/1511.01644}.

\bibitem[Liu et~al.(2008)Liu, Ting, and Zhou]{liu:08}
Liu, F.~T., Ting, K.~M., and Zhou, Z.-H.
\newblock Isolation forest.
\newblock In \emph{IEEE ICDM}, 2008.

\bibitem[Macha \& Akoglu(2018)Macha and Akoglu]{macha:2018}
Macha, M. and Akoglu, L.
\newblock Explaining anomalies in groups with characterizing subspace rules.
\newblock \emph{Data Mining and Knowledge Discovery}, 32\penalty0 (5):\penalty0
  1444--1480, 2018.

\bibitem[Mu{\~n}oz-Mar{\'\i} et~al.(2010)Mu{\~n}oz-Mar{\'\i}, Bovolo,
  G{\'o}mez-Chova, Bruzzone, and Camp-Valls]{munoz2010semisupervised}
Mu{\~n}oz-Mar{\'\i}, J., Bovolo, F., G{\'o}mez-Chova, L., Bruzzone, L., and
  Camp-Valls, G.
\newblock Semisupervised one-class support vector machines for classification
  of remote sensing data.
\newblock \emph{IEEE transactions on geoscience and remote sensing},
  48\penalty0 (8):\penalty0 3188--3197, 2010.

\bibitem[Nissim et~al.(2014)Nissim, Cohen, Moskovitch, Shabtai, Edry, Bar-Ad,
  and Elovici]{nissim:14}
Nissim, N., Cohen, A., Moskovitch, R., Shabtai, A., Edry, M., Bar-Ad, O., and
  Elovici, Y.
\newblock Alpd: Active learning framework for enhancing the detection of
  malicious pdf files.
\newblock In \emph{IEEE Joint Intelligence and Security Informatics
  Conference}, 2014.

\bibitem[Pevny(2015)]{pevny:2015}
Pevny, T.
\newblock Loda: Lightweight on-line detector of anomalies.
\newblock \emph{Machine Learning}, 102\penalty0 (2):\penalty0 275--304, 2015.

\bibitem[Ribeiro et~al.(2016)Ribeiro, Singh, and Guestrin]{ribeiro:2016}
Ribeiro, M.~T., Singh, S., and Guestrin, C.
\newblock ``why should {I} trust you?'': Explaining the predictions of any
  classifier.
\newblock In \emph{Proceedings of the 22nd {ACM} {SIGKDD} International
  Conference on Knowledge Discovery and Data Mining {(KDD)}}, pp.\  1135--1144,
  2016.

\bibitem[Ribeiro et~al.(2018)Ribeiro, Singh, and Guestrin]{ribeiro:2018}
Ribeiro, M.~T., Singh, S., and Guestrin, C.
\newblock Anchors: High-precision model-agnostic explanations.
\newblock In \emph{Proceedings of the Thirty-Second {AAAI} Conference on
  Artificial Intelligence {(AAAI-18)}}, pp.\  1527--1535, 2018.

\bibitem[Sch{\"o}lkopf et~al.(2001)Sch{\"o}lkopf, Platt, Shawe-Taylor, Smola,
  and Williamson]{scholkopf2001estimating}
Sch{\"o}lkopf, B., Platt, J.~C., Shawe-Taylor, J., Smola, A.~J., and
  Williamson, R.~C.
\newblock Estimating the support of a high-dimensional distribution.
\newblock \emph{Neural computation}, 13\penalty0 (7):\penalty0 1443--1471,
  2001.

\bibitem[Senator et~al.(2013)Senator, Goldberg, Memory, Young, Rees, Pierce,
  Huang, Reardon, Bader, Chow, Essa, Jones, Bettadapura, Chau, Green, Kaya,
  Zakrzewska, Briscoe, Mappus, McColl, Weiss, Dietterich, Fern, Wong, Das,
  Emmott, Irvine, Lee, Koutra, Faloutsos, Corkill, Friedland, Gentzel, and
  Jensen]{senator:2013}
Senator, T.~E., Goldberg, H.~G., Memory, A., Young, W.~T., Rees, B., Pierce,
  R., Huang, D., Reardon, M., Bader, D.~A., Chow, E., Essa, I., Jones, J.,
  Bettadapura, V., Chau, D.~H., Green, O., Kaya, O., Zakrzewska, A., Briscoe,
  E., Mappus, R. I.~L., McColl, R., Weiss, L., Dietterich, T.~G., Fern, A.,
  Wong, W.-K., Das, S., Emmott, A., Irvine, J., Lee, J.-Y., Koutra, D.,
  Faloutsos, C., Corkill, D., Friedland, L., Gentzel, A., and Jensen, D.
\newblock Detecting insider threats in a real corporate database of computer
  usage activity.
\newblock In \emph{KDD}, 2013.

\bibitem[Siddiqui et~al.(2018)Siddiqui, Fern, Dietterich, Wright, Theriault,
  and Archer]{siddiqui:2018}
Siddiqui, M.~A., Fern, A., Dietterich, T.~G., Wright, R., Theriault, A., and
  Archer, D.~W.
\newblock Feedback-guided anomaly discovery via online optimization.
\newblock In \emph{Proceedings of the 24th {ACM} {SIGKDD} International
  Conference on Knowledge Discovery {\&} Data Mining, {(KDD)}}, pp.\
  2200--2209, 2018.

\bibitem[Stokes et~al.(2008)Stokes, Platt, Kravis, and Shilman]{stokes:2008}
Stokes, J.~W., Platt, J.~C., Kravis, J., and Shilman, M.
\newblock Aladin: Active learning of anomalies to detect intrusions.
\newblock \emph{Technique Report. Microsoft Network Security Redmond, WA},
  98052, 2008.

\bibitem[Veeramachaneni et~al.(2016)Veeramachaneni, Arnaldo, Cuesta-Infante,
  Korrapati, Bassias, and Li]{veeramachaneni:2016}
Veeramachaneni, K., Arnaldo, I., Cuesta-Infante, A., Korrapati, V., Bassias,
  C., and Li, K.
\newblock Ai2: Training a big data machine to defend.
\newblock \emph{IEEE International Conference on Big Data Security}, 2016.

\bibitem[Zhu(2005)]{zhu2005semi}
Zhu, X.~J.
\newblock Semi-supervised learning literature survey.
\newblock Technical report, University of Wisconsin-Madison Department of
  Computer Sciences, 2005.

\end{thebibliography}
\bibliographystyle{icml2020}
\end{document}